\documentclass{article}

\usepackage[preprint]{neurips_2019}
\usepackage{natbib}
\setcitestyle{numbers}

\usepackage[utf8]{inputenc} 
\usepackage[T1]{fontenc}    

\usepackage{url}            
\usepackage{nicefrac}       
\usepackage{bbm}
\usepackage{latexsym}
\usepackage{graphicx}
\usepackage{color}
\usepackage{adjustbox}
\usepackage{microtype}
\usepackage{graphicx}
\usepackage{subfigure}
\usepackage{amsmath,bm}
\usepackage{amsfonts}
\usepackage{amssymb}
\usepackage[ruled,vlined]{algorithm2e}
\usepackage{booktabs}
\usepackage{hyperref}
\hypersetup{
    colorlinks=true,
    linkcolor=blue,
    citecolor=blue,
    urlcolor=blue
}
\usepackage{xspace}
\usepackage{cleveref}
\usepackage{multirow}
\usepackage{adjustbox}
\usepackage{authblk}
\setcitestyle{square}    

\newcommand{\mname}{{Clinical XLNet}\xspace}

\usepackage[shortlabels]{enumitem}

\title{\mname: Modeling Sequential Clinical Notes and Predicting Prolonged Mechanical Ventilation}
\author[1]{\large{\textbf{Kexin Huang*}}} 
\author[2]{\textbf{Abhishek Singh*}} 
\author[3]{\textbf{Sitong Chen*}}
\author[4]{\textbf{Edward T. Moseley}}
\author[4]{\textbf{Chih-Ying Deng}}
\author[5]{\textbf{Naomi George}}  
\author[4,5]{\textbf{Charlotta Lindvall}} 
\affil[1]{\small{Health Data Science, Harvard T.H. Chan School of Public Health}}
\affil[2]{Media Lab, Massachusetts Institute of Technology}
\affil[3]{Biomedical Informatics, Harvard Medical School}
\affil[4]{Dana-Farber Cancer Institute}
\affil[5]{Brigham and Women's Hospital}
\affil[]{\texttt{kexinhuang@hsph.harvard.edu, abhi24@mit.edu, sitong\_chen@hms.harvard.edu, nrgeorge@bwh.harvard.edu, \{edward\_moseley,chih-ying\_deng,charlotta\_lindvall\}@dfci.harvard.edu}}

\begin{document}

\maketitle

\begin{abstract}
Clinical notes contain rich data, which is unexploited in predictive modeling compared to structured data.  In this work, we developed a new text representation \mname for clinical notes which also leverages the temporal information of the sequence of the notes. We evaluated our models on prolonged mechanical ventilation prediction problem and our experiments demonstrated that \mname outperforms the best baselines consistently.
\end{abstract}

\section{Introduction}

Unstructured clinical notes within Electronic Health Records (EHR) contain valuable information to support clinical decisions~\cite{murdoch_detsky_2013}. However, most prognostic models used in medical practice currently rely on scoring systems that only incorporates structured data~\cite{apache_1986,gall_1993,sofa_1996,rapsang_shyam_2014}. Yet to make accurate clinical decisions, clinicians have to go through numerous clinical notes to extract additional prognostic information from unstructured text. This adds to the workload of clinicians when they need to make quick decisions and thus may introduce human errors. Therefore, predictive models that utilize unstructured data could be very helpful in medical practice.

A major challenge to unleash the clinical power of unstructured data is in representing notes in ways that allow effective mining of clinically meaningful knowledge. Natural Language Processing (NLP) methods can be used to generate effective notes representation and exploit the predictive values. There are many recent advances in the standard NLP domain, such as BERT~\cite{Devlin2019BERTPO}, XLNet~\cite{Yang2019XLNetGA}. However, clinical notes are far different from the general domain text (Wikipedia, BookCorpus, etc). For example, clinical notes contain jargon and abbreviations, different grammar and syntax. It is notoriously difficult to obtain an effective note representation. Hence, to bridge the gap, it requires to design an architecture that captures the nuances of the individual clinical note while taking into account the temporal nature of the sequence of the notes. Recently, ClinicalBERT, which adapts the BERT model from the standard NLP domain to model clinical notes~\cite{clinicalbert, Alsentzer:2019} achieved superior performance in clinical text prediction. High-quality relationships between human-interpreted medical concepts in clinical notes were uncovered and ClinicalBERT outperformed competitive baselines in predicting 30-day hospital readmission~\cite{clinicalbert}. However, previous works still have the following limitations:

\begin{enumerate}[topsep=0pt, partopsep=0pt, itemsep=2pt,parsep=2pt]
    \item \textbf{Notes representation could be improved.} In the standard NLP domain, BERT ignores the discrepancy of masked positions between the pretraining and finetuning stage. An autoregressive pretraining method named XLNet has been proposed, which empirically outperforms BERT by a large margin on many NLP tasks \cite{Yang2019XLNetGA}. XLNet overcomes the pretrain-finetune discrepancy by using the permutations of factorization sequences to capture the bidirectional context. This presents an opportunity to further improve clinical note representation by adapting XLNet into the clinical domain.
    
    \item \textbf{Failure to incorporate the temporal dimension of clinical notes.} Clinical notes have a temporal dimension where the order of information in sequential notes can provide additional predictive signals. Many previous models~\cite{clinicalbert, Alsentzer:2019} only aggregates individual risk scores from each note which ignore the temporal information charted in EHR.
    
    \item \textbf{Unrealistic prediction setup from clinical reality.} Many previous works \cite{cnnrp_2017,ehr_HFSA_2015,wang_zhang_he_zha_2018} have used predictors that would not be available at the time point a clinical decision must be made. 
\end{enumerate}

In this paper, we present \mname, which processes a patient's notes and predict the probability of PMV and mortality. In particular, this model mitigates the aforementioned limitations via the following technical contributions:
\begin{enumerate}[topsep=0pt, partopsep=0pt, itemsep=2pt,parsep=2pt]
    \item \textbf{Improved clinical notes representation.} We apply the permutation language modeling method proposed in XLNet on the corpus from clinical notes to generate better clinical embeddings, as demonstrated in Section.~\ref{sec:result}.
    
    \item \textbf{Inclusion of temporal information.} We maintain the temporal order of the note embeddings generated from \mname and feed them into a bidirectional LSTM layer~\cite{Hochreiter1997LongSM}, which leverages information along the temporal dimension (Fig.~\ref{fig:method}).
    
    \item \textbf{Realistic prediction setup.} We performed meticulous cohort curation in MIMIC-III dataset \cite{mimic} with the clinician team and set up an actionable prediction task according to the real clinical setting. 
    Clinical notes used in this prediction task are strictly within the 48-hour time window.
\end{enumerate}

To evaluate \mname, we examined its performance to predict prolonged mechanical ventilation (PMV). PMV consumes a substantial amount of healthcare resources, results in great financial and emotional burdens for patients and their families, and is associated with high one-year mortality around 50-60\% \cite{mcgee_2010,nelson_carson_bice_2015,unroe_2010}. It is projected that over $600,000$ patients in the United States will require PMV by the year 2020 \cite{zilberberg_luippold_sulsky_shorr_2008}. A surgical procedure to create an opening (stoma) in the trachea, named tracheotomy, allows breathing through an alternative airway  \cite{cox_carson_holmes_howard_carey_2004}. Tracheotomy results in improved patient comfort, decreased duration of ICU and hospital stay, and reduced mortality \cite{mallick_bodenham_2010}. 

However, the problem with tracheotomy, as an invasive procedure, is that it may not be necessary if a patient's condition improves quickly. Thus, an early and correct decision of tracheotomy is critical. The clinical team make this decision based on available ICU evidence as well as clinical judgment on the likelihood of PMV and potential mortality. The current prognosis of these factors relies on the ProVent score, which incorporates only a limited number of structured data in the EHR \cite{udeh_hadder_udeh_2015}.

We compared \mname with several state-of-the-art baselines including BERT \cite{Devlin2019BERTPO}, XLNet \cite{Yang2019XLNetGA}, and ClinicalBERT \cite{clinicalbert} on both PMV and mortality predictions. Experimental results show \mname outperformed the best baselines consistently (Section.~\ref{sec:result}).

\section{Data}

We use the Multiparameter Intelligent Monitoring in Intensive Care III (MIMIC-III)  dataset \cite{mimic} hosted on PhysioNet \cite{physionet} for our model development and experiment. It consists of 58,976 unique admissions from 38,597 patients in the intensive care unit of the Beth
Israel Deaconess Medical Center (BIDMC) between 2001 and 2012 and it has 2,083,180 clinical note events. The data in MIMIC-III has been de-identified, and the institutional review boards of the Massachusetts Institute of Technology (No. 0403000206) and BIDMC (2001-P-001699/14) both approved the use of the database for research. 

\paragraph{Cohort Selection.} Comprehensive inclusion and exclusion criteria are applied onto the MIMIC-III dataset to generate our patient cohort who were above 18 years old, and were on mechanical ventilation for at least 2 days with more than 6 hours each day. The patient was also not organ donors and was nor transferred patients from other hospitals. To alleviate confounding, we further remove patients with diseases that always lead to PMV such as neuromuscular disease, malignant neoplasm, extensive burns, etc.

For each admission, we use the first ICU stay. For clinical notes, we are interested in physician and nursing notes, but we narrow down to nursing and respiratory notes within 48 hours from the start of the first ventilation event. The reason for only selecting nursing-related notes is to expand the cohort as MIMIC-III is missing physician notes from 2001 to 2008. Additional criteria that are applied in our data curation process are in Fig.~\ref{fig:flowchart}. In the end, we obtain a cohort of 7,287 unique patients and their corresponding 73,224 clinical notes. Table.~\ref{tab:tableOne} shows the cohort demographics.

\paragraph{Cohort Labels.} Our cohort is labeled with PMV and 90-day mortality in binary. PMV is defined as being on mechanical ventilation for more than 7 days with at least 6 hours each day \cite{PMV:2007}. For 90-day mortality, it is defined as death occurring within 90 days of first ICU admission.

\begin{table}[t]
    \centering    
    \caption{Cohort Statistics. For continuous variable, it reports mean with the standard deviation. For categorical variable, the count is given with percentage. }
    \begin{tabular}{l|c|cc|cc}
    \toprule
    Statistics & \textbf{All} & \textbf{MV $\geq$ 7d} & \textbf{MV < 7d} & \textbf{Survive < 90d} & \textbf{Survive $\geq$ 90d} \\ \hline
    Admissions & 7,287 & 3,412 & 3,875 & 2,680 & 4,607 \\ 
    Age & 64.3 (16.7) & 63.8 (16.6) & 64.8 (16.9) & 69.3 (15.4) & 61.4 (16.8) \\
    Male & 4,072 (55.8) & 1,936 (56.7) & 2,136 (55.1) & 1,474 (55.0) & 2598 (56.4)\\ \hline
    Ethnicity & & & & &  \\ 
    \hspace{6pt} White & 5,159 (70.8)& 2,431 (71.2) & 2,728 (70.4) & 1,844 (68.8) & 3,315 (72.0)\\ 
    \hspace{6pt} Black & \phantom{0,}590\phantom{0} (8.1) & \phantom{0,}259\phantom{0} (7.6)& \phantom{0,}331 \phantom{0}(8.5)& \phantom{0,}202\phantom{0} (7.5)& \phantom{0,}388 \phantom{0}(8.4)\\
    \hspace{6pt} Hispanic/Latino & \phantom{0,}215 \phantom{0}(2.9)& \phantom{0,0}89 \phantom{0}(2.6)& \phantom{0,}126 \phantom{0}(3.3)& \phantom{0,0}51 \phantom{0}(1.9)& \phantom{0,}164 \phantom{0}(3.6)\\ 
    \hspace{6pt} Asian & \phantom{0,}150 \phantom{0}(2.1)& \phantom{0,0}65 \phantom{0}(1.9)& \phantom{0,0}85 \phantom{0}(2.2)& \phantom{0,0}54 \phantom{0}(2.0)& \phantom{0,0}96 \phantom{0}(2.1)\\ 
    \hspace{6pt} Others & 1,173 (16.1)& \phantom{0,}568 (16.6) & \phantom{0,}902 (23.3)& \phantom{0,}529 (19.7)& \phantom{0,}644 (14.0) \\ \hline
    Notes & & & & &  \\ 
    \hspace{6pt} Word Count & 1774 (1645) & 1745 (1610) & 1799 (1674) & 1811 (1730) & 1753 (1593) \\ 
    \hspace{6pt} Note Count & 9.78 (4.70) & 9.54 (4.51)& 10.0 (4.86) & 9.72 (4.70)&  9.82 (4.70) \\ 
    \bottomrule
    \end{tabular}
    \label{tab:tableOne}
\end{table}

\section{Methods}
\begin{figure}[t]
    \centering
    \includegraphics[width = \textwidth]{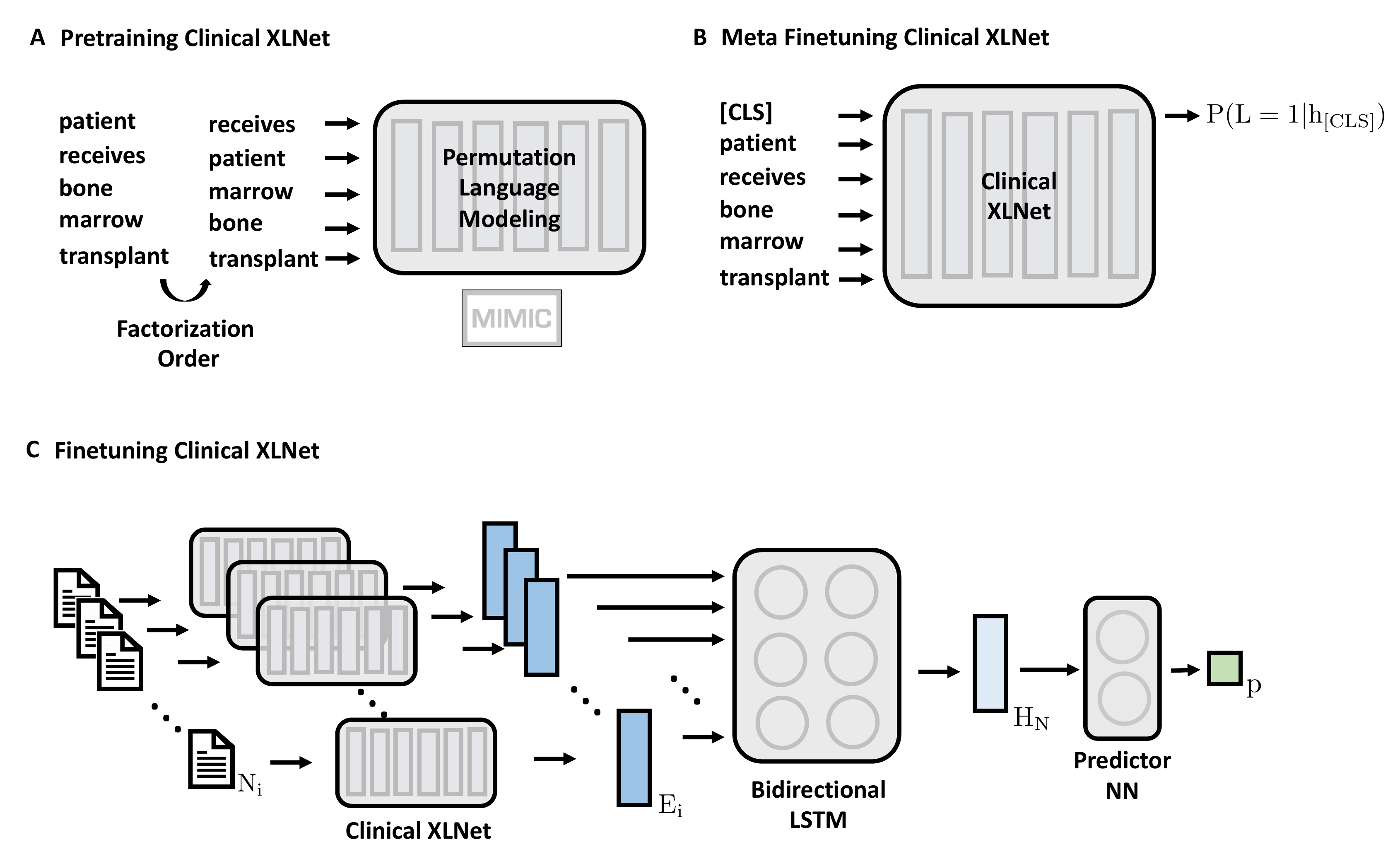}
    \caption{\mname framework. A. We first pre-train the XLNet embedding with MIMIC-III clinical notes dataset using Permutation Language Modeling. After pre-training, given a clinical note, the model outputs a numerical vector to be used as a note representation. B. To alleviate the computation burden from training end-to-end, the meta-finetuning stage uses the supervised signal to further tune the pre-trained network with input consisting of individual note $\mathrm{N_i}$. The meta finetuned stage then generates a static task-specific note representation. C. Given a sequence of a patient's notes $\mathrm{\{\mathrm{N_1}, \cdots, \mathrm{N_i}\}}$, the meta-finetuned \mname network generates a sequence of representation of notes $\mathrm{\{\mathrm{E_1}, \cdots, \mathrm{E_i}\}}$. The ordered representation sequence is then fed into a bidirectional LSTM layer, which then outputs a fixed size latent vector $\mathrm{H_N}$, representing the entire sequence. $\mathrm{H_N}$ is finally fed into a predictor neural network to generate a probability $\mathrm{p}$ measuring the likelihood of the target variable.}
    \label{fig:method}
\end{figure}

This section presents our \mname framework (Fig.~\ref{fig:method}). \mname is an extension of XLNet~\cite{Yang2019XLNetGA} on the clinical text domain. It first generates a deep latent representation for clinical notes and then by applying a bidirectional LSTM (Bi-LSTM) layer, it also leverages the sequential order of notes. 

\paragraph{Problem Settings.}\label{sec:prob} Our target task aims to leverage a patient's clinical notes to predict a patient's prognostic variables such as PMV and mortality. We denote a patient as $\mathrm{P}$, and each patient $\mathrm{P}$ is associated with a ordered sequence of notes $\{\mathrm{N_1}, \cdots, \mathrm{N_i}\}$, where $\mathrm{i}$ is the total number of notes. 

To predict mortality $\mathrm{L_M}$ and PMV $\mathrm{L_P}$, we aim to learn two mappings $\mathcal{M}: \{\mathrm{N_1}, \cdots, \mathrm{N_i}\} \longrightarrow [0, 1]$ and $\mathcal{P}: \{\mathrm{N_1}, \cdots, \mathrm{N_i}\} \longrightarrow [0, 1]$ where $[0 ,1]$ is a probability that measures the likelihood of having mortality and PMV respectively. 

\paragraph{Pretraining \mname.}\label{sec:notes} The text representation generated from large pre-training models depends on the corpus it is pre-trained on. XLNet is pre-trained on common language corpora such as BookCorpus, Wikipedia, Common Crawl and etc. However, these corpora are different from clinical notes which are filled with jargon, abbreviations and difficult syntax and grammar. Hence, to learn an effective representation of clinical notes, we need to further pre-train the XLNet using clinical notes. Specifically, we use our interested notes type (nursing, nursing/others and respiratory therapy) in MIMIC-III dataset. The clinical notes we used in pre-training are NOT in the holdout test set to avoid biased results. 

XLNet is a stack of Transformer-XL encoder~\cite{Dai2019TransformerXLAL}. For pre-training, it uses Permutation Language Modeling (PLM) to tackle the challenge of [MASK] token information gap between pre-training and fine-tuning in BERT~\cite{Devlin2019BERTPO}. For each sequence, XLNet and BERT are both appended with a [CLS] classification token at the beginning of the sequence for downstream task usage. PLM first factorizes the input sequence into a list of order-factorized sequences. Then, it applies language modeling to predict the next word given previous words. For a more detailed description, we refer the readers to the original paper~\cite{Yang2019XLNetGA}. 

After pre-training, given any clinical note $\mathrm{N_i}$, we use the last encoder layer hidden representation $\mathrm{E_i}$ of the [CLS] token to represent the note. Note, as we train with a supervised signal in the downstream task, [CLS] token would gather useful information in the entire note sequence due to the Transformer-XL's self-attention mechanism. Now, given a temporally ordered sequence of notes associated with a patient, we obtain a temporally-ordered sequence of notes representations $\mathrm{\{\mathrm{E_1}, \cdots, \mathrm{E_i}\}}$. 

\paragraph{Finetuning \mname.}\label{sec:temp} To leverage the temporal information among the notes, we feed $\mathrm{\{\mathrm{E_1}, \cdots, \mathrm{E_i}\}}$ into a sequential modeling layer. Specifically, we use Bi-LSTM model~\cite{Graves2005FramewisePC,Hochreiter1997LongSM}. We use bidirectional model because not only the latter notes depend on the previous notes as patients develop their symptoms in a temporal order but also the latter notes may contain useful clinical knowledge to help enrich the representation of previous notes. The output of the Bi-LSTM layer $\mathrm{H_N}$ is then fed into a predictor neural network, which at last, generates a probability $\mathrm{p}$ that measures the likelihood of downstream target variable, PMV, and mortality. The network is then tuned using binary classification loss. 

\paragraph{Meta-Finetuning.}~\label{sec:comp} End-to-end training is ideal as it can further tune the note representation module to allow it fit the specific task at hand. However, as each patient is associated with a large number of notes, during training, the forward and backward information through the large note representation module networks for each note are all stored in GPU memory. This is not scalable. Instead, we propose to approximate the task-specific note representation through an additional meta finetuning stage. During meta finetuning, we use one piece of note as input, and further train the pre-trained \mname through the downstream task label signal from the corresponding patient. The meta finetuned network can then generate a task specific note representation. Then, during the finetuning stage, we freeze the note representation XLNet module, and use a static fixed note representation from the meta-finetuned XLNet. 

This meta finetuned stage also allows fast adaptation to new dataset for the same downstream task because we only need to train on the temporal Bi-LSTM layer, which is relatively small, and takes around 5 minutes to converge using one GPU. This is an ideal quality to have in a realistic ICU setup that requires fast adapted information.

\section{Experiment}\label{sec:result}
To evaluate our model, we propose to examine the prediction performance under realistic setup. We use the first 48 hours of clinical notes starting from the initial mechanical ventilation event to predict two variables: mechanical ventilation longer than 7 days and 90-days mortality. \footnote{Models and scripts are open sourced at \href{https://github.com/kexinhuang12345/clinicalXLNet}{https://github.com/kexinhuang12345/clinicalXLNet}.}

\paragraph{Hyperparameters.} For data split, we first obtain a 10\% holdout test set. Then we generate different 8:1 train:validation splits using different random seeds for model performance robustness examination. For the pre-training, we further pre-train the XLNet embedding for another 200K steps using 16 batch size. For the meta fine-tuning, we use 32 batch size with learning rate 1e-5 for four epochs with early stopping on the area under the receiver operating characteristic curve (AUROC) score of validation. For fine-tuning, we use a two layers Bi-LSTM module with batch size 128 and learning rate 1e-4. The pre-training and meta fine-tuning process was conducted on a server with 2 Intel Xeon E5-2670v2
2.5GHZ CPUs, 128GB RAM, and 2 NVIDIA Tesla P40 GPUs. 

\paragraph{Evaluation Metrics.} We use area under the receiver operating characteristic curve (AUROC) for evaluation, which are standard metrics in the clinical informatics domain.

\paragraph{Baselines.} We conduct a thorough set of experiments with some of the popular baselines: 

\begin{enumerate}[topsep=0pt, partopsep=0pt, itemsep=2pt,parsep=2pt]
    \item \textbf{LSTM}~\cite{Hochreiter1997LongSM} is the classic language modeling method that uses long term document memory.
    \item \textbf{LSTM + Attention} adds an attention layer on top of the sequence output of LSTM hidden layers. 
    \item \textbf{Hierarchical Attention Networks (HAN)}~\cite{Hochreiter1997LongSM} is a hierarchical LSTM designed specifically for document level text classification. 
    \item \textbf{Recurrent Convolutional Neural Network (RCNN)}~\cite{yang2016hierarchical} uses a recurrent structure on the classic CNN network to capture contextual information as far as possible. 
    \item \textbf{BERT}~\cite{Devlin2019BERTPO} uses transformer based bidirectional contextual representation through a massive pre-training dataset and is later finetuned using a task-specific signal. 
    \item \textbf{XLNet}~\cite{Yang2019XLNetGA} uses similar pre-training and fine-tuning fashion as BERT. However, XLNet uses permutation language modeling with TransformerXL network backbone.
    \item \textbf{ClinicalBERT}~\cite{clinicalbert, Alsentzer:2019} further pre-trains on BERT using MIMIC-III notes dataset.
    \item \textbf{\mname-mean} is an ablation study that uses conventionally used heuristic method to report the aggregated output from a sequence of notes, instead of the bidirectional LSTM layers.
\end{enumerate}
Note that for BERT, XLNet, and ClinicalBERT, we all attach a bidirectional-LSTM layer on top of them to leverage the sequential dimension of notes. And for ClinicalBERT, we pre-train using the same corpus as the \mname. These steps ensure a fair comparison between note representation power.

\paragraph{Results.} Table.~\ref{tab:my_label} reports the result for our prolonged mechanical ventilation and 90-days mortality tasks. \mname achieves the best results with AUROC score of 0.663 ($\pm$ 0.011) and 0.779 ($\pm$ 0.006) for PMV and 90-days mortality respectively. From the difference between clinically pre-trained embedding \mname \&  Clinical BERT and no pre-trained model BERT \& XLNet, we demonstrate the necessity of pre-training on domain-specific corpus. From the difference between Clinical BERT and \mname, we show our \mname has better note representation. From the difference between \mname and \mname-mean, we see the usage of sequential modeling of the temporal dimension of notes.

\begin{table}[t]
    \centering    
    \caption{Prediction result with three independent data splits mean and standard deviation reported. The left panel is the PMV prediction result and the right panel is the 90 days mortality prediction results.}

    \begin{adjustbox}{width=0.4\textwidth}

    \begin{tabular}{lc}
    \toprule
    Method & AUROC \\ \hline
    LSTM & 0.613 $\pm$ 0.006\\ 
    LSTM + Attention & 0.604 $\pm$ 0.009 \\
    HAN & 0.606 $\pm$ 0.007\\
    RCNN & 0.620 $\pm$ 0.003 \\
    BERT & 0.616 $\pm$ 0.022\\
    XLNet & 0.611 $\pm$ 0.007\\
    ClinicalBERT & 0.648 $\pm$ 0.011\\
    \textbf{\mname} & \textbf{0.663 $\pm$ 0.011}\\\hline
    \mname - mean & 0.656 $\pm$ 0.003 \\
    \bottomrule
    \end{tabular}
    \end{adjustbox}
    \quad
    \begin{adjustbox}{width = 0.4\textwidth}
    \begin{tabular}{lc}
    \toprule
    Method & AUROC \\ \hline
    LSTM & 0.590 $\pm$ 0.034\\ 
    LSTM + Attention & 0.743 $\pm$ 0.007 \\
    HAN & 0.715 $\pm$ 0.013\\
    RCNN & 0.744 $\pm$ 0.010 \\
    BERT & 0.734 $\pm$ 0.047\\
    XLNet & 0.664 $\pm$ 0.017 \\
    ClinicalBERT & 0.774 $\pm$ 0.006\\
    \textbf{\mname} & \textbf{0.779 $\pm$ 0.006}\\\hline
    \mname - mean & 0.773 $\pm$ 0.003 \\
    \bottomrule
    \end{tabular}
    \end{adjustbox}
\label{tab:my_label}
\end{table}

\section{Discussion}

In this work, we propose a method for predicting prognosis based on only contextual information available from clinician notes. The proposed method is based on the recent advancements in the field of NLP. Therefore we compare it with other recently proposed methods in NLP for a fair comparison. We perform one ablation study to show the relevance of the proposed sequential modeling of the notes embedding in the time domain and we compare against several state of the art baselines which have been used extensively in the natural language domain as well as in the clinical context. 

\paragraph{Clinical Relevance.}
Our work provides timely aid in clinical decision making. For a patient under the ICU observation, the clinical team could start the evaluation to consider a tracheotomy procedure as soon as 48 hours after mechanical ventilation. The time period of 48 hours was chosen in consultation with a team of clinicians. Furthermore, a predictive analytics on the prolonged mechanical ventilation for seven days or more is important for clinicians in deciding the tracheotomy decision. Besides, the doctors could reduce the risk of a burdensome procedure and treatment by considering the patient's 90-days mortality prediction. This approach assist patients and their families by providing more time to process and make a major decision.

\paragraph{Model Efficiency.}
The proposed method uses XLNet~\cite{Yang2019XLNetGA} which uses TransformerXL~\cite{Dai2019TransformerXLAL} as the base architecture to extract embedding from the notes. Since every set of notes require their individual embeddings, we run the base architecture for multiple runs where the number of runs is proportional to the number of notes. Therefore, obtaining the embedding for the whole sequence of notes is computationally expensive both during the training as well as inference. However, there is a recent line of work~\cite{lan2019albert} which can allow executing the transformer based models at a much lesser computational cost.

\paragraph{Limitations and Future Work.}
Our proposed method only mines task relevant information from clinical notes from nurses and respiratory therapists. However, one can utilize other sources of data as well such as structured notes. While structured data are commonly used in prognostic models, our preliminary study showed that they did not improve the performance by any significant factor. One future direction could be to explore a novel architecture design that could utilize both sources of information to improve the performance further. Another future direction would be to explore ways of combining multiple sources of clinical notes such as physician notes, admission notes, and dsischarge notes.

Another limitation is our \mname model was pre-trained on MIMIC-III de-identified notes. All the protected health information (PHI) in MIMIC-III dataset was removed and replaced with surrogates, which is a drawback if the downstream task is named entity recognition (NER). Future work should consider using synthetic identification of the clinical text as source data~\cite{Synthetic}.

\section*{Acknowledgement}
The project was conceived, designed and conducted during the 2019 fall course HST.953 Collaborative Data Science in Medicine at the Harvard-MIT Division of Health Science and Technology. We express our thanks to all HST faculty for support and guidance during this project.

\bibliography{refs}
\bibliographystyle{unsrtnat}
\clearpage
\appendix

\begin{figure}
    \centering
    \includegraphics[width=0.85\textwidth]{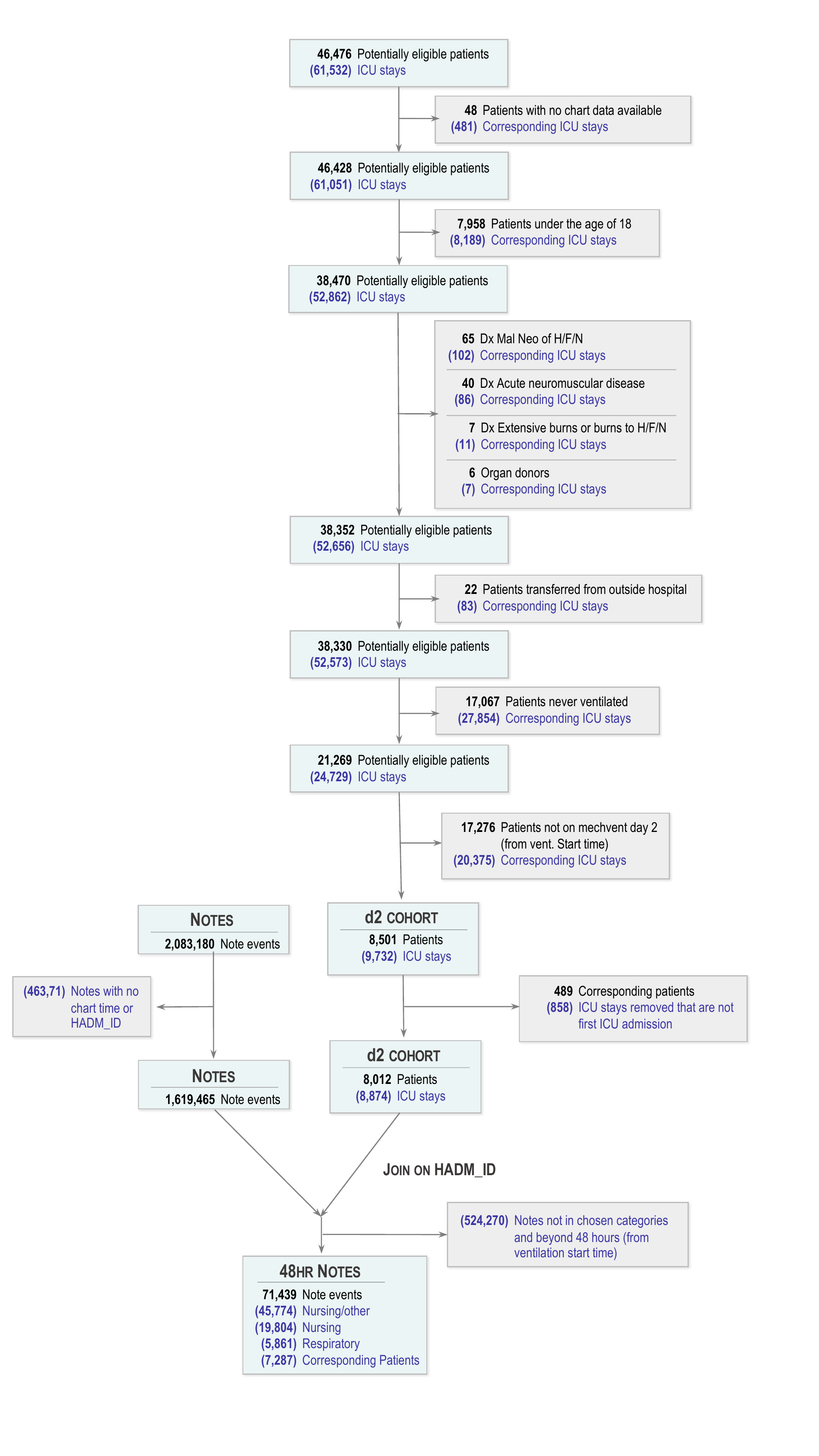}
    \caption{Cohort curation procedures in flow diagram. }
    \label{fig:flowchart}
\end{figure}

\end{document}